# Towards Credit-Fraud Detection via Sparsely Varying Gaussian Approximations

Harshit Sharma
*Department of Electrical Engineering*
*Indian Institute of Technology, Jodhpur*
Jodhpur, 342037–Rajasthan, India
sharma.23@iitj.ac.in

Harsh K. Gandhi
*Department of Electrical Engineering*
*Indian Institute of Technology, Jodhpur*
Jodhpur, 342037–Rajasthan, India
gandhi.1@iitj.ac.in

Apoorv Jain
*Department of Electrical Engineering*
*Indian Institute of Technology, Jodhpur*
Jodhpur, 342037–Rajasthan, India
jain.14@iitj.ac.in

*Abstract*—Fraudulent activities are an expensive problem for many financial institutions, costing billions of dollars to corporations annually. More commonly occurring activities in this regards are credit card frauds. In this context, credit card fraud detection concept has been developed over the lines of incorporating the uncertainty in our prediction system to ensure better judgment in such a crucial task. We propose to use sparse Gaussian classification method to work with the large data-set and use the concept of pseudo or inducing inputs. We perform the same with different sets of kernels and different number of inducing data points to show the best accuracy was obtained with the selection of RBF kernel with a higher number of inducing points. Our approach was able to work over large financial data given the stochastic nature of our method employed and also good test accuracy with low variance over the prediction suggesting confidence and robustness in our model. Using the methodologies of Bayesian learning techniques with the incorporated inducing points phenomenon, are successfully able to obtain a healthy accuracy and a high confidence score.

## I. INTRODUCTION

With the advent of digitalization of monetary funds, online transactions have played a vital role in the development of economies, and increase the quality of life around the world. These online transactions have become indispensable tool to the survival of a healthy day to day business proceedings of a small scaled business to big business corporations. One of the forms of these digital payment methods being credit cards which constitute of majority share when it comes to the pool of electronic payments that are wired daily. Due to amid dependence of the populations on such a handy and powerful tool to satisfy their needs of payments etc., for which they have become one of the most targeted hosts towards fraudulent activities. These fraudulent activities thereby cost corporations and governments billions of dollars and hence pose a vital problem yet to be solved in today's world of digital payments.

In that direction, there have been several attempts in order to visualize the problem of credit fraud detection so as to rectify it at the preliminary stages thus saving the financial institutions from the huge losses incurred. Many fraud detection techniques in fact aim to detect fraud accurately and even before fraud is committed. One of the techniques is to develop a model using decision tree along with the reinforcement using the Hunt and Luhn algorithms in order to detect the credit fraud transactions [1]. Another approach in this direction was proposed by utilizing the tools of data mining, visual cryptography and decision trees specifically designed to detect frauds [2]. In particular, the goal of these methods is to detect least and accurate false fraud detection. Recently there have been several of these proposals towards implementing a robust credit card fraud detection that utilize the methodology of K-means Clustering [1] or Hidden Markov Model [2], or Group Method of Data Handling [3], etc.

Recently new and novel methodologies based on the Dempster Shafer Theory [4] have also been reported to solve this critical issue before the losses incurred become paramount. However, the advent of Bayesian learning [5] and Neural Networks along with the algorithms that have been developed in the recent years that avail the fundamental techniques of machine learning have proved to show promising results in this context. In fact, for its flexibility, a Bayesian Network model to detect cyber crimes in which an inference process wherein the decision tree approach was employed to verify the consistency of the Bayesian Network [6].

In an article by Maes et al., different machine learning techniques such as the Bayesian Belief Network (BBN) and Artificial Neural Network (ANN) have been applied to analyse the problem statement and the results that were achieved significantly highlighted correlation with the real-world financial data. An important observation they reported was that Bayesian Network has a shorter training period and yields better results concerning fraud detection but the process of fraud detection is faster in artificial Neural Networks[7]. This approach of combining the machine learning techniques highlighted a huge potential of cost effective measure to tackle this issue of fraud detection. However, in a report published, Mukhanov had discussed the problems related to Bayesian Belief Network (BBN) in fraud detection. Towards compensating for this deficiency, development of an input data representation methods were even considered. Which led to the conclusion that the Naive Bayesian Classifier that is based on the input data representation method is accurate and but the underlying Bayesian Networks that were considered for analysis and experimentation were found to be accurate and user-friendly than a Naive Bayesian Classifier[8]. Further analysis towards proving that Bayesian networks could have an enhancement in performance in terms of detection of credit fraud has been reported by Dr. S. Geetha et al. [9]

With the basic of understanding that the Bayesian learning have already been reported, In this paper we seek to take the advantages of the flexibility and the versatility of the Gaussian processes that could be taken for the Bayesian learning that require very less data sufficiency to enable online learning to counter the real time predictive analysis towards credit fraud detection. In this direction, we employ the concept of inducing points and clustering to the sample from a large pool of unbalanced data wherein a clear minority and majority of the class of data could be demarcated. Referring to the fraudulent inclined data to be the minority, we use these tools that could assist the learning model to pickup the representation of the minority cases with an enhanced performance. During this process, we look at the optimized leveraging of the inducing points concept to provide minority fraudulent class representation so that the model can be self-equipped to perform predictive analysis for the future cases. We also lay down the comparison between the various kernels that can be employed for the purpose and provide the justification for the selection of the RBF kernel for the process. The overall proposal of this analysis is to execute the development of a system in the long run with real-time implementation scheme that incorporates the measure of certainty with which the prediction has been made to make sure that any case with low confidence of credit non-fraud is not being termed to safe case while the reality maybe opposite. The system should also be able to take advantage of online learning and scalable techniques for the implementation of Gaussian processes.

Further the paper is divided into 3 sections. In section 2, we exploit the methodology of Bayesian learning and lay down the mathematical understanding of how the framework of the model has been explained. In section 3, the involved performed experimentation of the model to test the accuracy and the performance has been explained to indicate the impact of the model performed. Finally, in section 4, the conclusion and further prospectives are given.

## II. METHODOLOGIES

In this section we try to illustrate the mathematical model and its analytical understanding of how the implementation of our model has been carried out for the experiment proposed to highlight the features of this model for its potential for execution in real-time problems. For this, we base the entire discussion of our methodology on the basic understanding of a Gaussian process. We extrapolate the properties and exploit the flexibilities of a Gaussian process for developing the mathematical model of the learning being executed in the paper. Using the basis that a Gaussian process [10] is a probability distribution (Gaussian) over functions, we employ this stochastic algorithm which forms a powerful tool as it exhibits a collection of random variables whose linear combination is a normal scatter forming a multivariate Gaussian distribution, thereby favoring it to be a very efficient tool through out the process of learning. This makes the Gaussian processes favorable for many applications in the domain of statistical machine learning, where it can be exploited for this strength by extracting the joint distributions.

For an appropriate demonstration, let us Consider $n$ observation values from an arbitrary dataset, $\mathbf{y} = \{y_1, \ldots, y_n\}$. This dataset can then be consider to resemble a single point obtained after sampling an ($n$-variate) Gaussian distribution. Therefore, we can now associate a dataset with a Gaussian process (GP). With this basis of how the dataset can be associated with a GP, we move towards judiciously setting the parameters i.e, mean and variance of the Gaussian process that can be applied in this methodology. Generally, it is assumed that the mean of this GP is zero for the sake of simplicity and but the main subject of interest comes from the co-variance function. For a general GP, The co-variance function $k(x, x')$ resembles the relation between the observations that follow the underlying GP is realized. The relation maybe defined as per the prior belief. Below is an example of such kernel based on squared exponential behaviour that is chosen for the sake of understanding and for its ability to be represented in other forms of relations that are defined in the literature as illustrated in Eq.1.

$$k\left(x, x'\right) = s_f^2 \exp\left[\frac{-\left(x - x'\right)^2}{2l^2}\right] \quad (1)$$

Here $x$ and $x'$ represent two arbitrary input points sampled from the dataset under consideration, $s_f^2$ represents the signal variance and $l$ represents the length-scale, which is a hyper-parameter for the kernel. The above formulation suggests that the training data has to be considered completely at the time of inference where the nature of inrference was non-parametric. Consequently, this makes GPs computationally expensive to work on with exact implementation scaling as $\mathcal{O}\left(N^3\right)$ time, and memory, where $N$ is the size of the training set. However, with the advent of sparse approximations [5], we shall be working with a lower computational cost, typically $\mathcal{O}\left(NM^2\right)$ time and $\mathcal{O}\left(NM\right)$ memory for some chosen $M < N$. All such approximations admit directional inference on fewer number of quantities, that represent around the whole posterior over functions.

Since we are interested in the prospect of utilization of Gaussian processes for classification, the GP is passed or can be said to be 'squashed' through a sigmoid inverse-link function. Also, towards our purpose that is to perform a classification task, a Bernoulli likelihood is considered to condition the data on the modified function values. Towards this we take more insight on the implementation of this Bernoulli scheme as mentioned in the reference of Rasmussen and Williams [10] for a better and robust understanding. With the motivation derived, we use the topological approach that has been expressed in the the graph below that summarizes the pipeline we plan to establish in order to perform the classification task with the proposed method.

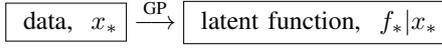

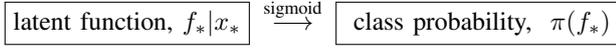

Furthermore, we collect the input data (say $N$ observations) into matrix $\{x_n\}_{n=1}^{N}$ and denote the binary class observations as $\{y_n\}_{n=1}^{N}$. To better understand we evaluate the co-variance function as per Eq.1, while pairwise picking the input vectors to compute on the equation to give the co-variance matrix $\{K_{nn}\}$ in the usual way. Hence, we obtain a prior for the values of the GP function at the input points given as a normal probability distribution that we denoted as $p(\mathbf{f}) = \mathcal{N}(\mathbf{f}|\mathbf{0}, \mathbf{K}_{nn})$.

Also the probit inverse link function that is to be utilized was denoted as $\phi(x)$ which can be defined as $\phi(x) = \int_{-\infty}^{x} \mathcal{N}(a|0,1) \mathrm{d}a$. We utilize the Bernoulli distribution classically denoted as $\mathcal{B}(y_n|\phi(f_n)) = \phi(f_n)^{y_n}(1-\phi(f_n))^{1-y_n}$. Further, we exploit the joint distribution of data and latent variables that can be explained as:

$$p(\mathbf{y}, \mathbf{f}) = \prod_{n=1}^{N} \mathcal{B}(y_n|\phi(f_n)) \mathcal{N}(\mathbf{f}|\mathbf{0}, \mathbf{K}_{nn}) \qquad (2)$$

The posterior over function values $p(f|y)$ is the main prospect of interest from the above equation, which can be considered through an approximated approach. We also require a condition where we can approximate the marginal likelihood $p(y)$ in order to further optimize (or marginalize) parameters of the co-variance function. In this direction, we take guidance and direction from an assortment of approximation schemes have been proposed [19], but they all require $\mathcal{O}(N^3)$ computation which we however originally aim to reduce. This motivation to reduce the cost of computation is therefore the reason we explore the concept of sparse Gaussian processes for a scalable method to deal with highly complex data for its ability to undermine the overall time complexity and get a favorable and lower resultant computational cost.

*Sparse Gaussian Processes*

As already established that a Gaussian process is fully determined by its mean $m(x)$ and co-variance $k(x, x')$ functions, it is therefore pivotal to select these parameters judiciously and with a carefully selected optimized values. For which, we assume the mean to be zero, without loss of generality as mentioned before. In addition, the co-variance function determines properties of the functions and should depend on the type of prior belief we are willing to incorporate based on our estimated knowledge of the domain of the data processed. A finite collection of function values at inputs $x_i$ follows a Gaussian distribution $\mathcal{N}(f : 0, K_{ff})$, where $\{K_{ff}\}_{ij} = $ k$(x_i, x_j)$. We thereby model the function of interest $f()$ using a GP prior, and noisy observations at the input locations $\{X\} = x_{ii}$ are detected in the vector $y$ where $y$ represents the observations corresponding to the input dataset.

$$p(\mathbf{f}) = \mathcal{N}(\mathbf{f} : 0, K_{ff}) \qquad (3)$$

$$p(y|f) = \prod_{n=1}^{N} \mathcal{N}(y_n, f_n, \sigma_n^2) \qquad (4)$$

The major objective is to make our model robust. Alternatively, we experiment with various co-variance functions such as the radial basis function which can be expressed as $k(x, x') = s_f^2 exp(-1/2|x - x'/l^2)$, wherein our results would now rely upon the initial minority based clustered pseudo-points decaying distance from one another. The length-scale $L$, the signal variance $s_f^2$, and the noise variance $\sigma_n^2$ comprise the hyper-parameter $\theta$, and is explicitly suppressed in the notation.

To make predictions, we choose to follow the common approach of first determining $\theta$ by optimizing the marginal likelihood and then marginalising over the posterior of $f^*$ following the relations:

$$\theta^* = \underset{\theta}{\mathrm{argmax}}\, p(\mathbf{y}|\theta) \qquad (5)$$

$$p(y^*|\mathbf{y}) = \frac{p(y^*, \mathbf{y})}{p(\mathbf{y})} = \int p(y^*|f^*)\, p(f^*|\mathbf{f})\, p(\mathbf{f}|\mathbf{y})\, \mathrm{d}\mathbf{f}\, df^* \qquad (6)$$

Unfortunately, the cost of computing the marginal likelihood, the posterior and the predictive distribution scales as $\mathcal{O}(N^3)$ regardless of the closed-form gaussian structure. This progressively happens due to the inversion of $\{K_{ff} + \sigma_n^2 I\}$, which is impractical for many datasets that are considered in real-life situations. Thereby comes the motivation of using the concept of inducing point phenomenon. Here, we plan to leverage on inducing point methods [5] to deal with the computational complexity given the high dimentionality of data used. The method suggests augmentation of latent variables with additional input-output pairs $\mathbf{Z}, \mathbf{u}$, known as 'inducing inputs' and 'inducing variables' respectively. The joint distribution now takes the form after the simplication and the consideration of inducing points and is expressed as:

$$p(\mathbf{y}, \mathbf{f}, \mathbf{u}) = p(\mathbf{y}|\mathbf{f}) p(\mathbf{f}|\mathbf{u}) p(\mathbf{u}) \qquad (7)$$

Integration over $f$ is usually very complicated and is generally obtained via approximation in order to obtain computationally efficient inference. To obtain the popular Fully Independent Training Condtional method (FITC method) in the case of Gaussian likelihood, a factorization is enforced such that $p(\mathbf{y}|\mathbf{u}) \approx \prod_n p(y_n|\mathbf{u})$. Additionally, to get a variational approximation, the following inequality is used that can set the upper bound to the training dataset as:

$$\log p(\mathbf{y}|\mathbf{u}) \geq \mathbb{E}_{p(\mathbf{f}|\mathbf{u})}[\log p(\mathbf{y}|\mathbf{f})] \triangleq \log \tilde{p}(\mathbf{y}|\mathbf{u}) \qquad (8)$$

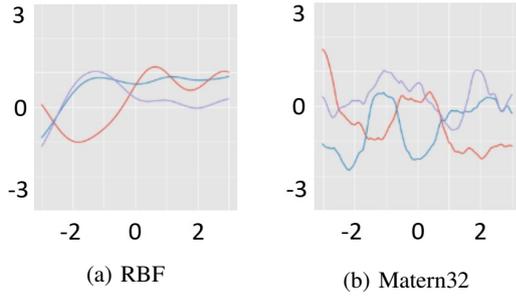

Fig. 1: Figure depicts the various kernels that have been used to test the model, in particular the comparison between RBF and Matern32 kernels. The random lines denote random possibilities/functions which are trained upon with data to converge to some points.

But To obtain a tractable bound on the marginal liklihood we substitute this bound on the conditional into the expression $p(\mathbf{y}) = \int p(\mathbf{y}|\mathbf{u})p(\mathbf{u})d\mathbf{u}$ as illustrated in Titsias, 2009 [11]:

$$\log p(\mathbf{y}) \geq \log \mathcal{N}\left(\mathbf{y}|\mathbf{0}, \mathbf{K}_{nm}\mathbf{K}_{mm}^{-1}\mathbf{K}_{nm}^{\top} + \sigma^2 \mathbf{I}\right) \\ - \frac{1}{2\sigma^2}\operatorname{tr}\left(\mathbf{K}_{nn} - \mathbf{Q}_{nn}\right) \quad (9)$$

where $\sigma^2$ represents the variance of the Normal likelihood term, $\{\mathbf{K}_{mm}\}$ is the covariance matrix function evaluated at all pairs of inducing inputs points $z_m, z_{m'}$, $\mathbf{K}_{nm}$ is the covariance function formed evaluating across the data input points and inducing inputs points and $\mathbf{Q}_{nn} = \mathbf{K}_{nm}\mathbf{K}_{mm}^{-1}\mathbf{K}_{nm}^{\top}$. The covariance function parameters can be then optimized using the bound obtained on the marginal likelihood.

Over the years, FITC has remained a popular inducing point method. However, in the case of Bernoulli liklihood, the required integrals for depicted in Eq.9 is not tractable. Generalized FITC method proposed by Naish-Guzman and Holden, 2007 deals with this problem. However, we shall be using the single variational bound method also known as the sparse $KL$ divegence method as demonstrated by Hensman [12] which has outperformed the Generalised FITC method. Other approaches such as sparse mean field approach also exist but these tend to give a result assimilating the laplace approximation which proves to be but of little help in classification task [12]. In the further sub-section, we show how the $KL$ divergence method is adapted in the model.

*Single variational bound*

To demonstrate the idea behind the selection of the single variational bound strategy we shall now build upon the traceable bound given in Eq.8 on the conditional used to construct the variational bounds for the Gaussian case as:

$$\log p(\mathbf{y}|\mathbf{u}) \geq \mathbb{E}_{p(\mathbf{f}|\mathbf{u})}[\log p(\mathbf{y}|\mathbf{f})] \quad (10)$$

which as discussed is in general intractable for the non-conjugate case (eg. the case we are dealing with i.e. Bernoulli).

In order To deal with this type of distributions we recall the standard variational equation:

$$\log p(\mathbf{y}) \geq \mathbb{E}_{q(\mathbf{u})}[\log p(\mathbf{y}|\mathbf{u})] - \operatorname{KL}[q(\mathbf{u})\|p(\mathbf{u})] \quad (11)$$

After some mathematical computations and substituting (10) into (11) results in a further bound on the marginal likelihood:

$$\begin{aligned}\log p(\mathbf{y}) &\geq \mathbb{E}_{q(\mathbf{u})}[\ log p(\mathbf{y}|\mathbf{u})] - \operatorname{KL}[q(\mathbf{u})\|p(\mathbf{u})] \\ &\geq \mathbb{E}_{q(\mathbf{u})}\left[\mathbb{E}_{p(\mathbf{f}|\mathbf{u})}[\log p(\mathbf{y}|\mathbf{f})]\right] - \operatorname{KL}[q(\mathbf{u})\|p(\mathbf{u})] \\ &= \mathbb{E}_{q(\mathbf{f})}[\log p(\mathbf{y}|\mathbf{f})] - \operatorname{KL}[q(\mathbf{u})\|p(\mathbf{u})]\end{aligned} \quad (12)$$

where for the sake of notation consider $q(\mathbf{f}) := \int p(\mathbf{f}|\mathbf{u})q(\mathbf{u})d\mathbf{u}$. Finally let us consider taking $q(\mathbf{u}) = \mathcal{N}(\mathbf{u}|\mathbf{m}, \mathbf{S})$, a variational distribution with parameters $\mathbf{m}, \mathbf{S}$ as prescribed in Hensman et al., 2013. This gives the following functional form for $q(\mathbf{f})$:

$$q(\mathbf{f}) = \mathcal{N}\left(\mathbf{f}|\mathbf{Am}, \mathbf{K}_{nn} + \mathbf{A}\left(\mathbf{S} - \mathbf{K}_{mm}\right)\mathbf{A}^{\top}\right) \quad (13)$$

where $\{\mathbf{A}\} = \{\mathbf{K_{nm}K_{mm}^{-1}}\}$. Since in the classification case the likelihood factors as $p(\mathbf{y}|\mathbf{f}) = \prod_{i=1}^{N} p(y_i|f_i)$, we only require the marginals of $q(\mathbf{f})$ in order to compute the expectations in eq.(12). Our algorithm then uses gradient based optimization to maximize the parameters of $q(\mathbf{u})$ with respect to the bound mentioned on the marginal likelihood. With the note-able robustness of the metric and the model that we define, we move on the experimentation and the learning with the subsequent results of the model to justify and verify the claim.

## III. EXPERIMENTATION AND RESULTS

**Data Information**: For the purpose of validation, the dataset has been collected and analysed during a research collaboration of the Machine Learning Group of ULB (Université Libre de Bruxelles) and Wordline on big data mining and fraud detection. The following dataset contains thirty distinct features which hold numerical values and weights that depict various parameters for the consideration of fraud or non-fraud cases. These values are a result of PCA that has been performed on the otherwise original data with larger number of features making them unfit for the usage with gaussian processes. Further data preprocessing was done on the features with very high value (such as Amount) by standard normalization scheme and the data was checked to do away with any trends and seasonality. The dataset is highly unbalanced, the positive class (frauds) account for 0.172% of all transactions.

Further, the experiment was performed with setup chosen as $M = 50$, for the start to check with the theoritical expectation. The initial setup of inducing points was formulated by clustering using $K$ means over an inverse dataset. This inverse dataset was formed with minorty class as majority and vice versa. The effects of increasing number of the inducing points was checked with the variation of accuracy and precision. Further, the test was conducted using various kernels which include the RBF (radial base function) as can be seen in Fig.1(a) and Matern32 kernel as can be referred to Fig.1(b) was also reported. The models prepared were tested for the effect of withholding the optimization over the inducing points and otherwise by attempting to study the tolerance study on the same.

Upon experimentation, the results obtained displayed that an increasing number of inducing points showed positively reinforced results on both the test accuracy and liklihood. The inducing points number selected were 50, 100, and 150. We also noticed an increase in the test accuracy with the increase in the number of inducing points and also a decrease in the test liklihood which portrays the confidence of the system over the test data-points. Overall, the performance obtained was best for $M = 150$ as in accordance with the analytical construction of the model. We also realized that modelling our prior belief about the data with a Radial basis fuction(RBF) kernel provided much better results than the combination of Matern32 and White kernels. The results have been formulated in the table 1 for reference.

The table has been formlated keeping in mind the Radial basis function that was provided with kernel parameters of length-scale ($l$)=1 without any active dimension bias and variance ($s_f^2$)=2. The input dimensions used were selected to be 30 in accordance with the dataset provided. Matern32 kernel mixture with white kernel was tested with white kernel variance parameter assigned as unity. The effect of withholding the pseudo points training was also realized to reveal null effect of increasing the number of inducing points with a sparse variational process. Since, our problem concerns classification task, a bernoulli liklihood assignment was common to all the processes carried out.

The sparse variational approximation performed exceptionally under the RBF kernel assumption to model the gaussian process multivariate variance with distance dependence of the input on the classification output. Other setups were proved less suited for our purpose with the output though being sensitive to the increasing inducing points number yet the overall performance was not the best obtained. Progressively for 50 inducing points the accuracy of proper and correct credit fraud detection for the RBF, Matern32 and the combination of Matern32 with the white kernel was found to be ≈ 97.9, 87.71, 88.01 out of the cases when 100 cases were taken. This accuracy rose gradually when the inducing points were now increased to 100 and 150 and their respective values were reported in the table 1. Also the likelihood of the testing for $M = 50$ for the three different setups with the different kernels were found to be -0.1132, -0.8016 and -0.7681 respectively.

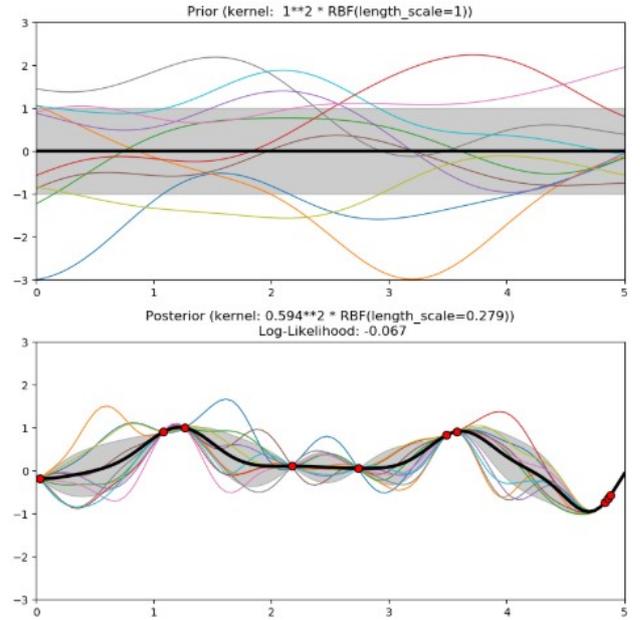

Fig. 2: Figure depicts the RBF kernel that has been used to test the model. The random lines denote random possibilities/functions which are trained upon with data to converge to some points.

The training was performed using mini batch stochastic optimization using Adam optimizer for gradient based optimization. Stochastic Gradient descent was also experimented without much change in the results. The dataset was distributed in the batches of hundred to perform the optimization. The results of such high accuracy with the subsequent minimal losses have been reported that can serve as a very novel credit-fraud detection technique.

## IV. CONCLUSION AND DISCUSSION

In summary, we first inspected the dataset through exploratory data analysis and reached upon the conclusion that the usage of smooth kernel priors in a Gaussian process should benefit the convergence to a sufficient solution for the problem. The nature of the variables provided been hidden, caused us to yet try a flexible prior settings and verifying our hypothesis. We, then inspected various methods to deal with the imbalance of the dataset provided to us and consequently dived into working with pseudo-points concept and sparse Gaussian processes to deal with the big volume data provided. Using the methodologies described based on the sparsely distributed gaussian processes incorporated with the phenomenon of inducing points, we were successfully able to obtain a healthy accuracy and confidence score.

|  | Inducing Points | | | | | |
| --- | --- | --- | --- | --- | --- | --- |
|  | M=50 | | M=100 | | M=150 | |
| kernels | Test Liklihood | Test Accuracy | Test Liklihood | Test Accuracy | Test Liklihood | Test Accuracy |
| **RBF** | **-0.1132** | **0.9790** | **-0.1047** | **0.9790** | **-0.1137** | **0.9805** |
| Matern32+ White | -0.8016 | 0.8771 | -0.7191 | 0.8801 | -0.7760 | 0.8831 |
| Matern32 | -0.7681 | 0.8801 | -0.7670 | 0.8801 | -0.7563 | 0.8816 |

TABLE I: Results that depict the accuracy and the likelihood for various inducing points used that equipped various kernels have been reported and have been tablulated.

It was found that using the strategy mentioned the dataset provided a test accuracy of 98.05% on a dataset with a skew of 15:85 percent ratio of fraudulent vs clean cases provided. We were even able to obtain a test liklihood of -0.1137 which is a measure of confidence. Lower magnitude provides a more confident prediction over the test cases passed. We realized that using large number of inducing points could help consider more diversity in data provided and improve the performance of the mentioned strategy. We also derived the inference that Radial basis function displaying the best result ability showcases the smoothness in the relation between the data points. Using the application of inducing points one can work with data having high class imbalance in the field of credit fraud detection given such nature of data. The model could be used to incorporate uncertainty in such crucial decision as that of credit fraud detection.